\documentclass[letterpaper, 10 pt, conference]{ieeeconf}  % Comment this line out if you need a4paper
\usepackage{setspace}
\usepackage{etoolbox}

\usepackage{afterpage}

\IEEEoverridecommandlockouts                              % This command is only needed if 
\AtBeginEnvironment{quote}{\singlespace\small}
\AtEndEnvironment{quote}{\vspace{-\topsep}\endsinglespace}

\usepackage{siunitx}
\usepackage{makecell}
\usepackage{array}
\usepackage{esvect}
\usepackage{booktabs,tabularx}
\usepackage{placeins}
\usepackage{algorithm}
\usepackage{algpseudocode}
\usepackage{amsmath}
\algtext*{EndProcedure}
\usepackage{cite}
\usepackage{amsmath,amssymb,amsfonts}
\usepackage{graphicx}
\usepackage{multirow}
\usepackage{subfig}
\usepackage{textcomp}
\usepackage{xcolor}
\usepackage[export]{adjustbox}
\usepackage{wrapfig}
\usepackage{diagbox}

\title{\LARGE \bf
Through the Clutter: Exploring the Impact of Complex Environments on the Legibility of Robot Motion
}

\author{Melanie Schmidt-Wolf$^{1}$, Tyler Becker$^{2}$, Denielle Oliva$^{3}$, Monica Nicolescu$^{4}$, and David Feil-Seifer$^{5}$ % <-this % stops a space
\thanks{$^{1}$Melanie Schmidt-Wolf, $^{2}$Tyler Becker, $^{3}$Denielle Oliva, $^{4}$Monica Nicolescu, and $^{5}$David Feil-Seifer are with the Department of Computer Science and Engineering, University of Nevada, Reno, 1664 N. Virginia Street, Reno, NV 89557-0171, USA \emph{mschmidtwolf@unr.edu}, \emph{tbecker@unr.edu}, \emph{denielleo@unr.edu}, \emph{monica@cse.unr.edu} and \emph{dave@cse.unr.edu}
}
\thanks{This material is based upon work supported under the AI Research Institutes program by National Science Foundation and the Institute of Education Sciences, U.S. Department of Education through Award \# 2229873 - AI Institute for Transforming Education for Children with Speech and Language Processing Challenges. Any opinions, findings and conclusions or recommendations expressed in this material are those of the author(s) and do not necessarily reflect the views of the National Science Foundation, the Institute of Education Sciences, or the U.S. Department of Education.
}
}

\begin{document}

\maketitle

\begin{abstract}

 The environments in which the collaboration of a robot would be the most helpful to a person are frequently uncontrolled and \textit{cluttered} with many objects present. Legible robot arm motion is crucial in tasks like these in order to avoid possible collisions, improve the workflow and help ensure the safety of the person. Prior work in this area, however, focuses on solutions that are tested only in \textit{uncluttered} environments and there are not many results taken from \textit{cluttered} environments.
% We propose a novel algorithm for legible motion planning based on potential fields.
In this research we present a \textit{measure for clutteredness} based on an entropic measure of the environment,  and a \textit{novel motion planner} based on potential fields. Both our measures and the planner were tested in a \textit{cluttered} environment meant to represent a more typical tool sorting task for which the person would collaborate with a robot.
The in-person validation study with Baxter robots shows a significant improvement in legibility of our proposed legible motion planner compared to the current state-of-the-art legible motion planner in \textit{cluttered} environments. Further, the results show a significant difference in the performance of the planners in \textit{cluttered} and \textit{uncluttered} environments, and the need to further explore legible motion in \textit{cluttered} environments.
% The in-person validation study with Baxter robots shows that our proposed legible motion planner leads to a marginally significant increase in legibility compared to the current state-of-the-art legible motion planner. 
We argue that the inconsistency of our results in \textit{cluttered} environments with those obtained from \textit{uncluttered} environments points out several important issues with the current research performed in the area of legible motion planners.

\end{abstract}

% \begin{keywords}

% \end{keywords}

\section{Introduction}\label{intro}

In human-robot collaboration tasks it is inefficient, frustrating, and a safety hazard when the intended goal of the robot is unclear. When people work together to complete a task they communicate both explicitly (e.g., verbally) and implicitly (e.g., through arm motions) in order to show their intention. This way they are able to avoid collisions and overall increase the efficiency with which they can complete the task. For robot collaborators the task is harder, as expectations from the human for the robot are higher overall than when working with another person. Furthermore, the patience when this collaboration fails is much smaller.

This problem becomes even more difficult in \textit{cluttered} environments because the effectiveness of both explicit and implicit communication decreases due to the increased number of objects in the scene and as well due to their close proximity to each other. Furthermore, as the sensitivity of the environment increases so does the importance of preventing collisions. In this paper, we investigate legible motion in \textit{cluttered} environments which focuses on the ability of the robot to express its intent implicitly through the trajectory of its arm while completing a task.

\begin{figure}
    \centering
    \captionsetup[subfigure]{justification=centering}
        \subfloat[][One of the Baxter robots with an \textit{uncluttered} environment setup]{\includegraphics[width=0.24\textwidth]{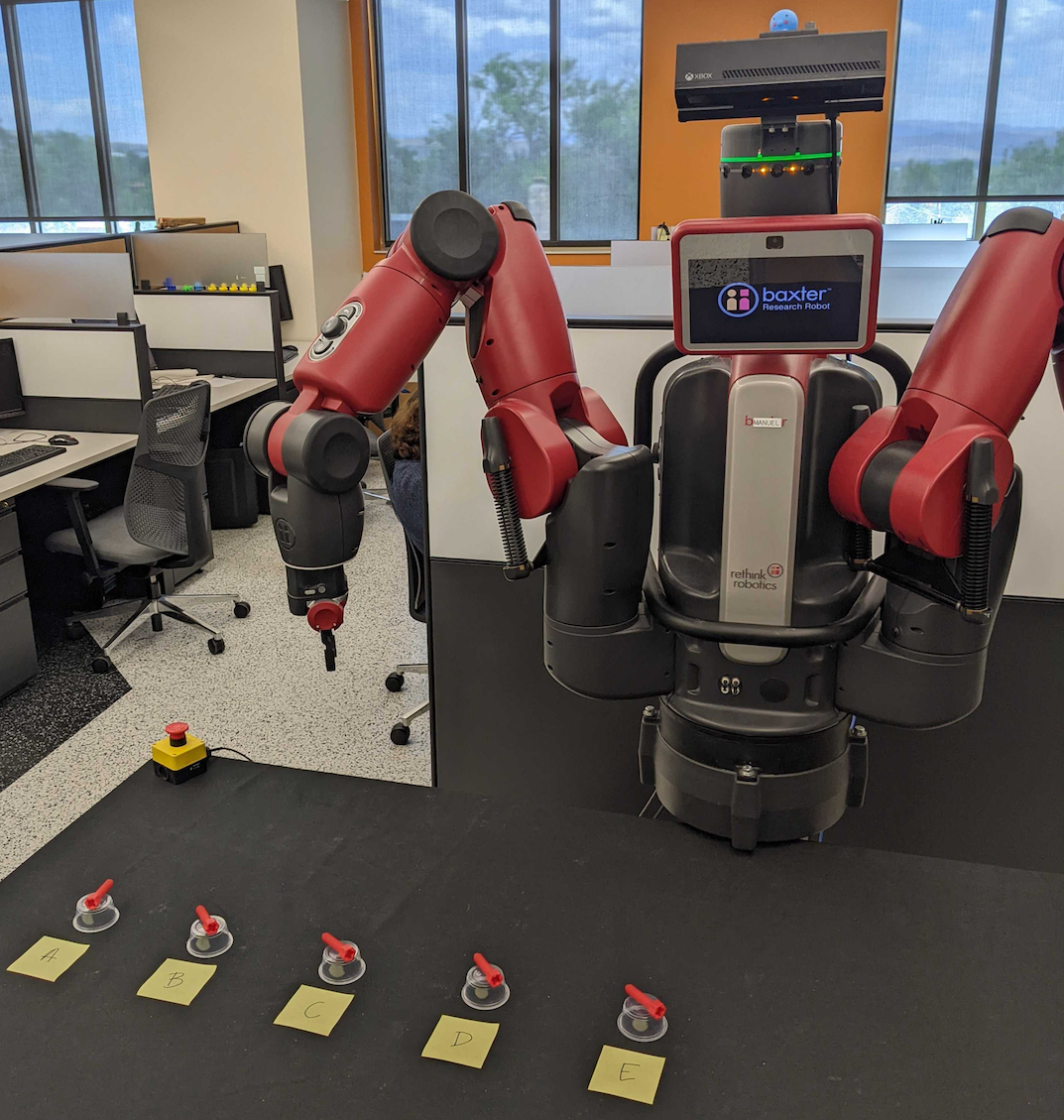}\label{baxter1}} 
        \subfloat[][Another Baxter robot with a \textit{cluttered} environment setup]{\includegraphics[width=0.24\textwidth]{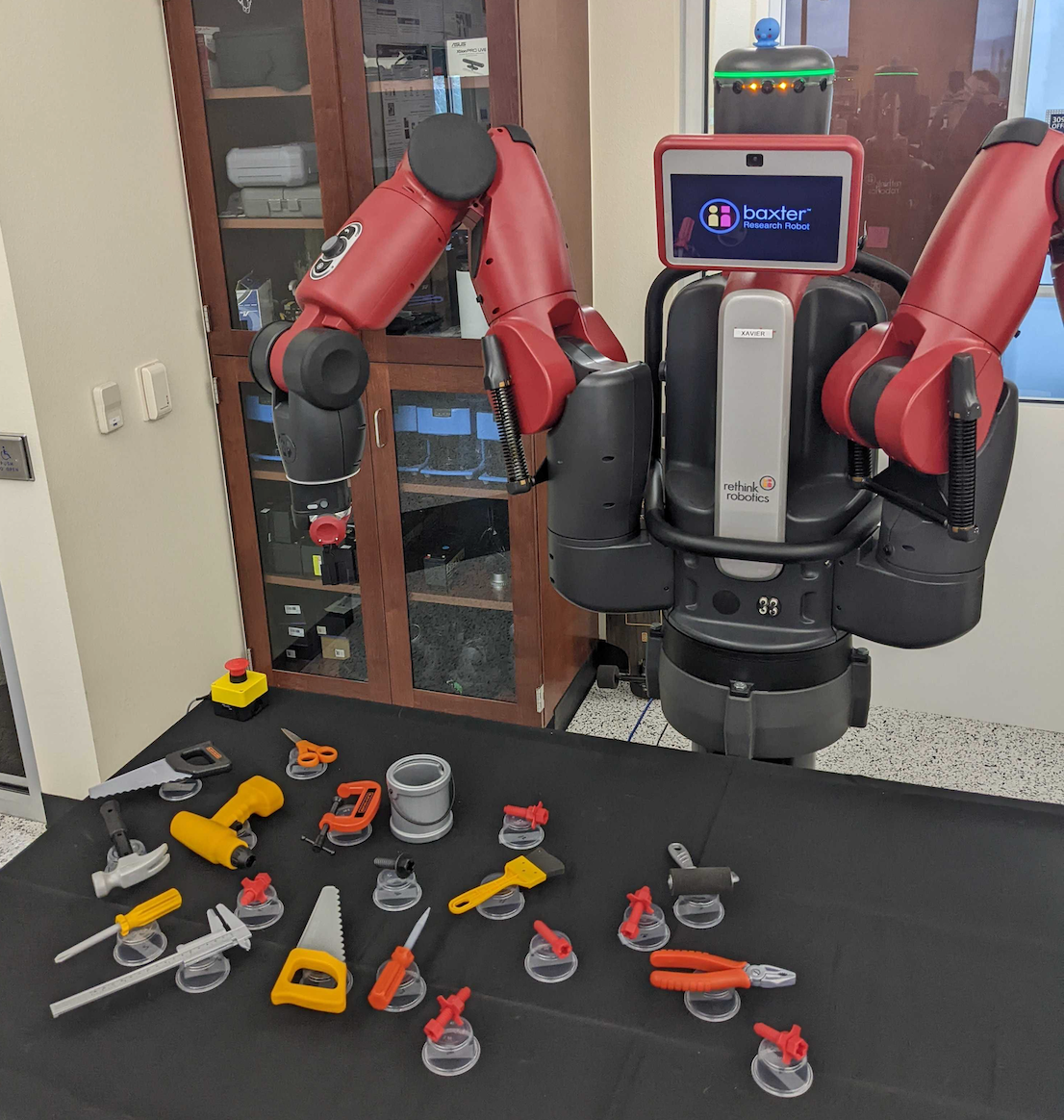}\label{baxter2}} 
    \caption{Baxter humanoid robots with the two environment setups used in the validation study.}    
    \label{fig:baxter_robots}
    % \vspace{-1em}
\end{figure}

Previous solutions developed for legible motion have modeled it as an optimization problem for which the constraints are defined using a variety of methods. Previously proposed constraints include: 
% relative position to the goal; where the end effector is pointing; the velocity of the end effector; the overall distance of the trajectory; dissimilarity to other trajectories; and linearity of the path. 
\begin{itemize}
\item relative position to the goal;
\item where the end effector is pointing;
\item the velocity of the end effector;
\item the overall distance of the trajectory;
\item dissimilarity to other trajectories; and
\item linearity of the path. 
\end{itemize}
The results of these different methods were shown experimentally through user studies, however, the user studies were designed using simple \textit{uncluttered} environments.

While useful for initial studies, \textit{uncluttered} environments are not the norm for tasks where robot collaboration would be useful (e.g., cleaning tasks or sorting tasks). Therefore, in this work we test the previous state-of-the-art as well as a logical extension of that method in both an \textit{uncluttered} and a \textit{cluttered} environment (shown in Fig. \ref{fig:baxter_robots}) in order to determine both what effect a \textit{cluttered} environment has on the accuracy with which the intention can be determined and the effectiveness of the previous optimization criteria.

% The robot's intent is difficult to determine if many objects are located in an environment. In human-robot collaboration, humans and robots work on a task together, such as in an industrial setting with an assembly task. In a collaborative task, humans use implicit nonverbal communication to increase the efficiency of task performance and reduce the impact of errors from miscommunication \cite{admoni2016robot}. In this paper, we investigate the nonverbal communication form legible motion. 
% Legible motion is defined as motion that allows the human collaborator to infer the correct target object quickly and confidently \cite{dragan2013legibility}. Legible motion makes the robot's intentions clear and intent-expressive to the \mbox{human \cite{dragan2013legibility}}. 

The major contributions of this paper are as follows:
\begin{itemize}
\item A mathematical definition of the \textit{clutteredness} of a table environment based on the concept of entropy;
% \item A definition of the \textit{legibility} of a single object and show experimentally how it relates to the expected accuracy for a guessing collaborator.
\item A novel algorithm for generating legible motion trajectories based on the concept of entropy-scaled potential fields;
\item Demonstration of the impact of an environment's \textit{clutteredness} on the planner’s performance; and
\item An in-person user study using Baxter robots to evaluate the effectiveness of two different legible motion planners in both \textit{cluttered} and \textit{uncluttered} environments. 
\end{itemize}

The remainder of this paper is organized as follows: Section \ref{Background} covers the related work, Section \ref{Method} describes our approach to legible motion, Section \ref{Validation} describes the experiment design for the study, Section \ref{Results} reports on the results of the study, Section \ref{Discussion} provides our discussion of the results, and Section \ref{Conclusion} concludes the paper.
% In this paper, we investigate how the robot arm should move towards an object in a cluttered environment when collaborating with a human to avoid conflicts such as picking up the same object. 
% One of many challenges in human-robot collaboration is determining what actions the robot should take when collaborating with a human while avoiding conflicts such as picking up the same object. 
% To facilitate human-robot collaboration, we propose a novel algorithm for legible motion planning that is usable in cluttered environments. \\
% In this paper, we propose a novel algorithm for legible motion planning using potential fields for cluttered environments, and validated the approach with a study.
% We propose a novel algorithm for legible motion planning to facilitate human-robot collaboration.
% We propose a novel algorithm for legible motion planning that applies potential fields to avoid other objects that are not the target object to improve legible motion in cluttered environments. 
% Further, we validated the approach with a study. \mbox{Figure \ref{fig:baxter_robots}} shows the humanoid Baxter robots with the experiment setups that we used in the validation study.

% The proposed approach can enhance the safety of human-robot collaboration, and  facilitate collaboration of a robot and a human in a shared workspace with a cluttered environment.

\section{Related Work} \label{Background}

\subsection{Legible robot arm motion}

% Legible robot motion is clear to read, i.e. it is simple to understand the robot's intent \cite{wallkotter2022new}.

Legible robot motion implicitly expresses the intent of the robot to increase the understanding of a human collaborator. 
Further, legible robot motion is defined as motion that allows the human collaborator to infer the correct target object quickly and confidently \cite{dragan2013legibility}.
Prior work in this area has focused on developing motion planners which express the information in a way that people can understand more easily.  
Dragan et al. \cite{dragan2013legibility} developed mathematical models to define and distinguish predictability and legibility. This was evaluated in an \textit{uncluttered} environment with two objects using recorded videos.
Bied and Chetouani proposed a solution using reinforcement learning~\cite{observerIntegrated} to maximize their proposed metric for legibility. This was evaluated in an abstracted graphical environment and was not tested through a user study. As such, it was evaluated with a single object in an \textit{uncluttered} environment with a variable number of observers. Faria, et al., proposed a solution \cite{multipartlegiblemotion} for multiple people observing a motion which involved optimizing for the best value for them all rather than just a single person. This was evaluated in a \textit{uncluttered} environment using videos produced in simulation to show participants of a user study. Wallk{\"o}tter, et al., utilized supervised learning \cite{SLOTV} to generate legible motion by training on data that was evaluated and label through legibility measures that had been tested in prior works. The testing environment they used included seven unevenly spaced objects, and their results were validated through scoring accuracy and not through a user study. Most recently Bronars, et al., used conditional generative models guided by a legibility measures from previous work to generate legible motion\cite{bronars2023legible}. This system was evaluated through comparison to other planners by scoring their legibility with a measure, and they used an \textit{uncluttered} environment with two evenly spaced objects. 

A survey of ten legibility frameworks was provided by Wallk{\"o}tter, et al., \cite{wallkotter2022new} and they found that the legibility framework from Bodden, et al., performed the best~\cite{bodden2018flexible}. While we recommend this survey to get an understanding of work in this area and as well of the evaluation methods that are considered best, the planners in this survey were tested in a simulated \textit{uncluttered} environment with three objects present. 

In this research, we validate the chosen legible motion planners on real robots in a \textit{cluttered} and \textit{uncluttered} environment. 
The potential inconsistencies between \textit{cluttered} and \textit{uncluttered} environments may present significant challenges for existing approaches when employed in real-world environments.
% If \textit{cluttered} is not equal \textit{uncluttered} then realistic environments could pose a problem for existing approaches.
To address these challenges, we evaluate the use of potential fields as a possible solution for legible motion in \textit{cluttered} environments. We also evaluate this approach through an in-person user study with the motion planner on a real robot, and we investigate if clutter has an impact on the legibility of the motion planners.

% Wallkotter et al. \cite{wallkotter2022new} compared ten legibility frameworks with the result that the legibility framework from Bodden et al. \cite{bodden2018flexible} is currently the best-performing framework. 
% The comparison was made with a study that contained three objects \cite{wallkotter2022new}. For detailed information about the different legibility frameworks we refer to the review and comparison by Wallkotter et al. \cite{wallkotter2022new}.

% While the existing legibility motion planners perform better than a simple straight motion to the goal or as an explicit signal such as pointing \cite{bodden2018flexible}, these legibility motion planners were not designed to work for cluttered environments. We did not find research that investigated legible robot arm motion in cluttered environments.

\subsection{Cluttered Environment Measures}

% the need for cluttered env.
% how can it be measured
% different components

As illustrated in Fig. \ref{fig:baxter_robots}, the experimental validation will explore both an \textit{uncluttered} and a \textit{cluttered} environment. We expect that it might be more difficult for the robot arm to move legibly in a \textit{cluttered} environment.
In this work, we want to test and evaluate a robot's motion in a \textit{cluttered} environment. 
To achieve this, we need a measure for a \textit{cluttered} environment to consider the amount of \textit{clutteredness} in the shared workspace.

\textit{Clutteredness} is typically measured as a ratio of occupied and total space. This includes the number of obstacles affecting the robot, defined as the ratio of sensed obstacles to a preset \textit{clutter} value \cite{ram1997case}, as a measure proportional to voxel numbers from voxel occupancy grids \cite{lou2022learning}, or from a computer vision perspective as the sum of spectral residual values of the pixels and the total pixels in the candidate regions \cite{chun2013floor}. \textit{Clutter} can be defined as the distance between objects or if objects are in contact with each other by filtering through a point cloud of pixels~\cite{megha2012manipulation}. Further, entropy can be associated with disorder \cite{brissaud2005meanings}. In this paper, we establish a \textit{cluttered} environment measure from the Kullback-Leibler Divergence \cite{kullback1951information} which is based on entropy.

Other works \cite{Dogar2012}, \cite{muhayyuddin2018randomphys}, and \cite{nam2019occlusion}  propose methods for manipulating objects in \textit{cluttered} environments with the goal of finding valid paths, but not with the goal of legible motion in the case of human-robot interaction in a \textit{cluttered} space. The work in this paper investigates legible robot motion with the assumption of collaboration in a shared space.

\subsection{Potential Fields}

% how else are pot. fields used + examples

Artificial potential fields was proposed by Khatib \cite{khatib1986real} as a real-time obstacle avoidance approach. 
The artificial potential field approach bases the motions of the robot on how it would move if it were being affected continuously by artificial forces assigned to obstacles in the environment. The field consists of an attractive force to move towards the target and a repulsive force to avoid obstacles. Other work employed a potential field method for the design safe path planning in a collaborative task~\cite{chen2009assemly}. This method considered a hand of a human collaborator as an obstacle that affected the paths that could be taken in a dynamic assembly task. 

Although potential field algorithms are commonly used in robot path planning for obstacle avoidance, they have not been used to create legible motion.
In this paper, we apply potential fields with forces scaled by our \textit{clutteredness} measure to make the robot arm motion more legible.

\section{Approach} \label{Method}

In this section we describe our novel definition of the \textit{clutteredness} of objects in an environment, and a novel architecture for legible motion based on potential fields.

% In this section we describe our novel algorithm for legible motion planning based on collision free navigation using potential fields. First, we describe two new parameters that we use to scale the strength of each obstacle's potential field. Then, we describe our changes to how the repulsive and attractive forces are calculated to generate the trajectory. Finally, we describe the whole algorithm that we use to generate legible motion.

\subsection{Clutteredness Measure} In order to measure the \textit{clutteredness} of the environment we use the concept of entropy from information theory which gives an estimate of the surprise of receiving some given results produced by some distribution \cite{shannonEntropy}. We connect this to \textit{clutteredness} by comparing the distribution of the objects in the environment to an imagined uniform distribution of the same objects. For this measure, objects that are uniformly distributed are considered not \textit{cluttered}, and therefore we estimate the distribution of object positions as a Multivariate Gaussian and compare it to the uniform distribution using the \textit{Kullback-Leibler Divergence}\cite{klDivergence} as shown in Eq. \ref{KL_divergence}.

\begin{equation}\label{entropy}
h(p)=\mathbf{E}[-\log p(X)]=-\int_Xp(x)\log p(x)\mathbf{d}\textbf{\textit{x}}
\end{equation}
\begin{equation}\label{cross_entropy}
h(p,q)=\mathbf{E}_p[-\log q(X)]=-\int_Xp(x)\log q(x)\mathbf{d}\textbf{\textit{x}}
\end{equation}

\begin{equation}\label{KL_divergence}
D(p\|q)=-h(p)+h(p,q)
\end{equation}

In the equation, $p$ is the estimated Gaussian distribution, $q$ is the Uniform Distribution, $h(p)$ is the entropy, and $h(p,q)$ is the cross entropy of the two distributions. This measure of divergence, however, is unbounded and therefore not suitable as a scaling factor, so we use the non-linear normalized transformation shown in Eq. \ref{nln_transformation} to get our measure. 

\begin{equation}\label{nln_transformation}
\xi=e^{-D(p\|q)}
\end{equation}

This transformation guarantees that the value of $\xi$ is always positive and normalized.

\subsection{Legible Potential Fields}\label{LegibleMotionPotentialFields} 

The potential fields algorithm for robot navigation works by assigning a virtual repulsive force to obstacles in the environment and a virtual attractive force to the target. The robot is then pulled towards the target and pushed away from the obstacles and this generates a collision free path through the environment to the target. A general overview of the potential fields algorithm is given in \cite{khatib1986real}. 

\begin{figure}
% \vspace{-1em}
\centerline{\includegraphics[width=0.45\textwidth]{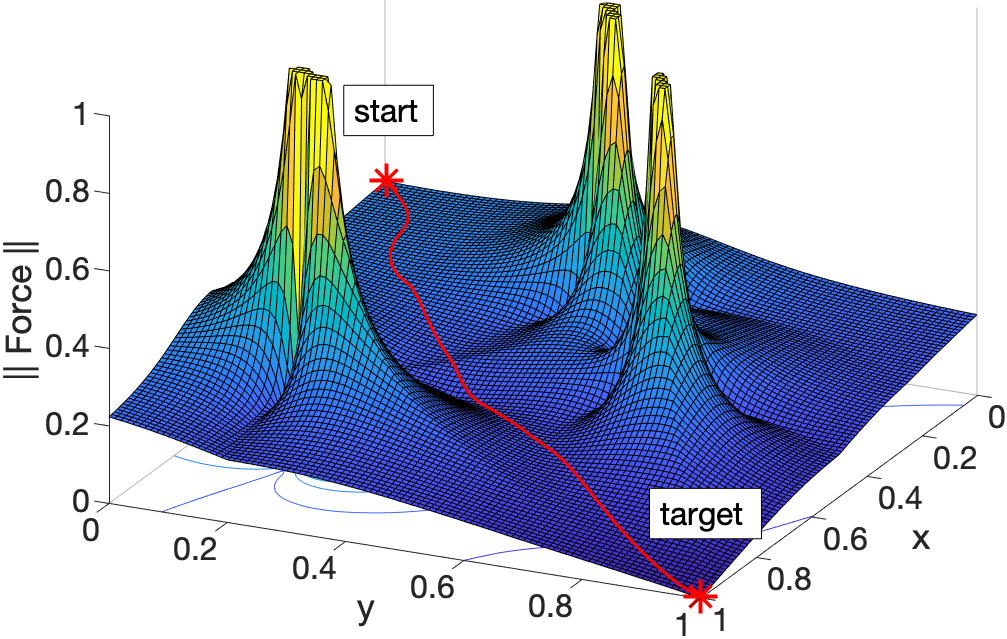}}
\caption{
Visual explanation of the path generated by the potential field, displayed with three obstacles.
}
\label{force}
% \vspace{-1em}
% \vspace{-0.5em}
\end{figure}

For this paper we applied this theory to trajectory planning in order to achieve legible motion in \textit{cluttered} environments. Our virtual attractive force is assigned to the target object and all other objects in the environment are assigned a virtual repulsive force which is scaled based on the measured $\xi$ value of the environment and a measure of how close the object is to a straight line trajectory. We then apply these forces to the position of the end effector until it reaches the target object, and we obtain a collision free trajectory that leads to the target. A smoother is also applied to the trajectory which both ensures that the path gives the obstacles a wider margin and that the trajectory only decreases in height. An example output of the potential fields algorithm is shown in Fig. \ref{force}. \par

Our assumption is that this will produce legible motion in \textit{cluttered} environments because this follows the optimization parameters described by Bodden et al. in \cite{bodden2018flexible}. They describe three parameters that they optimize for to achieve legible motion: \textit{Point Position}, \textit{Pointing}, and \textit{Velocity}. The results shown by Bodden et al. are not promising with respect to \textit{Pointing}, and therefore we have omitted this consideration from our planner as well as the implementation of their planner in the study whose results are shown in \mbox{Section \ref{Results}}. However, we will describe below how by calculating the trajectory with potential fields we also naturally follow both the \textit{Point Position} and \textit{Velocity} parameters in \textit{cluttered} environments.

\textit{Point Position} is described as a heuristic that predicts the goal of a trajectory based on which object is closest to the position of the end effector. By optimizing this value for the target object the motion planner generates trajectories, which stays higher up until the end effector is over the goal and then immediately drops down to the object as shown in \mbox{Fig. \ref{fig3}}. Potential fields produce similar trajectories because of the force in the z-direction which forces the trajectory to stay far above objects. The trajectories generated from potential fields are lower than the trajectories from the state-of-the-art planner; we expect this will help convey closeness to the target in the \textit{cluttered} environment.

\textit{Velocity} is described as a heuristic which rewards the end effector for moving faster when farther away from the target object. This is under the assumption that the person will try to extrapolate the target using this information and the distance from other objects. Potential fields naturally do this as the attractive force is greater the farther the end effector is from the target object resulting in a higher speed. The difference with the proposed planner to the state-of-the-art is that the generated trajectory will slow down in order to move around objects that are in the way. We think this will give more information to the person collaborating with the robot in the \textit{cluttered} environment. 

The results of testing both our planner and the state-of-the-art planner in our in person study are shown in \mbox{Section \ref{Results}}.

\subsection{Algorithm} 

For clarity, we include a description of our implemented potential fields algorithm for legible motion in Alg. \ref{pf_algo}. The algorithm applies the standard potential fields navigation algorithm with minor modifications. A difference with our algorithm is that the trajectory initially produced contains areas in which it abruptly goes up and over obstacle fields. We implement a post-algorithm smoother, which eliminates these motions and instead makes the height of the trajectory always decreasing. This is due to the fact that the abrupt upwards motions made it harder to understand the intent of the motion. In two dimensions, the trajectory remains unchanged. Further, we applied two scaling factors, as described in section \ref{LegibleMotionPotentialFields}.

\begin{algorithm}
    \caption {Returns a legible trajectory from the starting position of the end effector to the position of the target object. $\vec{x}_{ee}$ is the starting position of the end effector and $j$ is the index of the target object.}
    \begin{algorithmic}
        \Procedure{$genLegibleTraj$}{$\vec{x}_{ee}$, $j$}
        \State $\vec{x}_{plan} \gets \vec{x}_{ee},\ traj\gets[\ ]$
        \While{$\|\vec{x}_{plan} -  \vec{x}_j\| < \epsilon$}
            \State $append(traj, \vec{x}_{plan})$
        
            \State $\vec{x}_{plan} \gets \vec{x}_{plan} - k_{update}\nabla U_{total}(j)$
        \EndWhile

        \State $traj\gets smooth(traj)$

        \State \Return $traj$
        \EndProcedure
    \end{algorithmic}

    % \vspace{-1.5em}
    \label{pf_algo}
\end{algorithm}

\begin{figure}
\centerline{\includegraphics[width=0.45\textwidth]{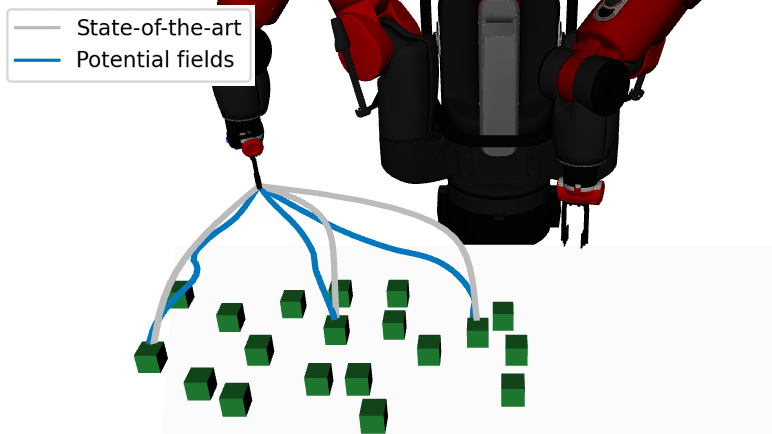}}
% \centerline{\includegraphics[width=0.3\textwidth]{Figures/Fig3Draft.png}}
\caption{
Comparisons of the trajectories of the state-of-the-art legible motion planner (gray) and our proposed legible motion planner based on potential fields (blue). Our entropy-scaled potential field legible motion planner avoids other objects (green), which leads to more legible motion.
}
\label{fig3}
% \vspace{-1em}
\end{figure}

\mbox{Fig. \ref{force}} shows an example of the potential field for three obstacles and a target object. The generated path from the start position of the end effector to the target avoids moving towards the obstacles and increases therefore legibility.
In \mbox{Fig. \ref{fig3}} a \textit{cluttered} environment setup with three example comparisons of the state-of-the-art and our proposed legible motion planner is visualized. Similar to \mbox{Fig. \ref{force}} our proposed trajectory avoids moving towards other objects in the shared workspace, which increases the trajectories' legibility.

\section{Experimental Validation} \label{Validation}
% To validate that the proposed legible motion planner generates more legible robotic grasping motions than established methods by using potential fields and entropy, and to evaluate if there are potential inconsistencies between cluttered and uncluttered environments that may present significant challenges for existing approaches when employed in real-world environments, we conducted a validation study with two Baxter robots (see \mbox{Fig. \ref{fig:baxter_robots}}).
We conducted a validation study with two Baxter robots (see \mbox{Fig. \ref{fig:baxter_robots}}) to verify if the proposed legible motion planner generates more legible robotic grasping motions than established methods by using potential fields and entropy. Additionally, we aimed to evaluate if there are  potential inconsistencies between \textit{cluttered} and \textit{uncluttered} environments that may pose challenges for existing approaches when applied in real-world environments.
40 participants were recruited via flyers and social media to participate in the
IRB-approved (IRBNet ID: 2090377-2) study. The user study had a duration
of about 30 minutes.

\subsection{Hypotheses}

% In this validation study, we want to evaluate if there are potential inconsistencies between cluttered and uncluttered environments that may present significant challenges for existing approaches when employed in real-world environments. We also want evaluate our proposed legible motion planner based on the concept of entropy-scaled potential fields.  

% Our hypotheses are as follows:
As part of the validation study, we are investigating the following hypotheses:

\begin{itemize}
\item {\bf H1:} Clutter has a negative impact on the legibility of the motion.
\end{itemize}

We will assess the validity of H1 by comparing participant responses between the \textit{uncluttered} and the \textit{cluttered} experiment setup by testing for significance to determine observed differences between both experiment setups.

\begin{itemize}
\item {\bf H2:} The proposed legible motion planner will result in more legible motion for: 
\begin{enumerate}
\item \textit{uncluttered} environments; and % (setup 1)
\item \textit{cluttered} environments % (setup 2)
\end{enumerate}
compared to the state-of-the-art legible motion planner from Bodden et al. \cite{bodden2018flexible}.
\end{itemize}

To validate H2, we will compare the participant's responses for both legible motion planners for the \textit{uncluttered} environment (hypothesis 2.1) and the \textit{cluttered} environment (hypothesis 2.2). The comparison will be conducted using statistical tests for significance to identify any significant differences between both legible motion planners.

\subsection{Study Design}
In this validation study, participants were asked to observe the robot as it picks up an object among many in a \textit{cluttered} environment. The trajectories were split into sections and after each section the participants were asked which object they think the robot is reaching for by ranking their choices from one to five, with rank 1 being their first choice. The participants answered questions about the executed motion and repeated this for different objects. 
Since every participant evaluates both the proposed legible motion planner and the state-of-the-art legible motion planner from Bodden et al. \cite{bodden2018flexible}, this study employs a within-subjects design.
The survey concluded with demographic questions.
For the study, we used two experiment setups. The first experiment setup place the objects as in the state-of-the-art paper with five objects next to each other, see \mbox{Fig. \ref{baxter1}}. The second experiment setup consists of a \textit{cluttered} environment with 20 objects, see \mbox{Fig. \ref{baxter2}}. The first experiment setup has a \textit{clutteredness} measure of $\xi$=1.0 and the second experiment setup has a \textit{clutteredness} measure $\xi$=0.06. In the validation study, the participants do not know which ten of the 20 objects the robot will approach. The order of the target objects and the motion planner were randomized.

% \subsubsection{Instruments}
% Baxter robot
% \subsubsection{Visualizations} 
% \subsubsection{Questionnaire} 
% \subsection{Participants}
% 30 participants

% \section{Quantitative Validation} \label{Score}
% Further, we established an illegibility score to validate our proposed legible motion planner quantitatively. For this, we counted the number of obstacles inside the projected surface of a field of view (FOV) on the plane, with the FOV origin at the end effector. To obtain the illegibility score, we divide the number of obstacles inside the projected surface by the total number of obstacles, see Figure \ref{fig:method} for a visual explanation example.

\begin{figure}[t]
    \begin{minipage}{0.91\linewidth}
    \centering  
\subfloat{\includegraphics[width=0.57\textwidth]{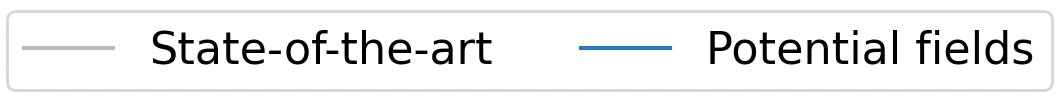}}    
% \subfloat{\includegraphics[width=0.7\textwidth]{Figures/LegendFig4.png}} 
    \end{minipage} \par\medskip
    \setcounter{subfigure}{0}
    \begin{minipage}{.49\linewidth}
    \centering
    \subfloat[][\textit{Uncluttered} setup: Section 1]{\includegraphics[width=1.0\textwidth]{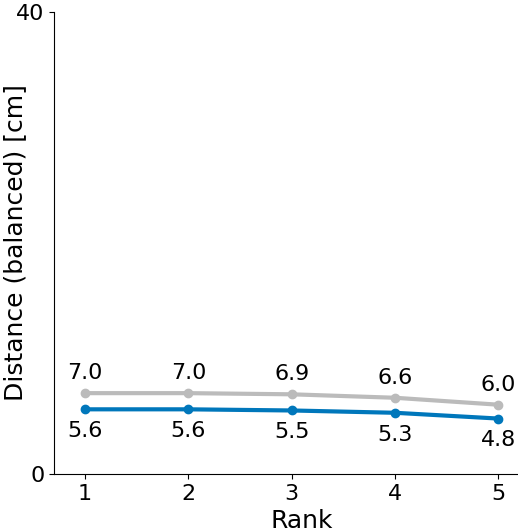}\label{section1_setup1}}
    \end{minipage}%
    \hfill
    \begin{minipage}{.49\linewidth}
    \centering
    \subfloat[][\textit{Uncluttered} setup: Section 2]{\includegraphics[width=1.0\textwidth]{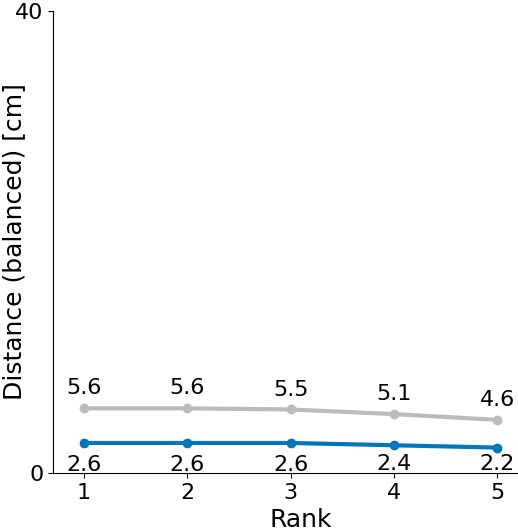}\label{section2_setup1}}
    \end{minipage}\par\medskip
    \caption{
    In an \textit{uncluttered} environment our entropy-scaled potential field legible motion planner and the state-of-the-art legible motion planner perform similarly well (\mbox{p-values $>$ 0.05}). In both sections of the trajectory, participants seem quite certain which object will be picked, especially when comparing the values with the values in \mbox{Fig. \ref{fig_rank}} (lower distances to the target object are better).
    }
    % \vspace{-1em}
    \label{fig_rank_setup1}
  \end{figure}

\section{Results} \label{Results}

Fig. \ref{fig_rank_setup1} shows the user study results for the \textit{uncluttered} environment. In \mbox{Fig. \ref{fig_rank}}, the comparison results of our entropy-scaled potential field planner and the state-of-the-art legible motion planner are visualized for the \textit{cluttered} environment. It is challenging to compare motion planners for objects in different parts of the environment because each object has a different distance and arrangement to other objects, and therefore the trajectories have a different legibility dependent on clutter. We compare the average distance to the correct object for the planner balanced by the average distance to other objects in the environment. 
The rank values in \mbox{Fig. \ref{fig_rank_setup1}} and \ref{fig_rank} are obtained by summing up the distances of the guessed object to the correct object until the participant guessed the correct object.

Since the Shapiro-Wilk test for normality achieved a p-value that is less than p $<$ 0.05 we cannot assume normality. Due to this result, we used non-parametric tests to calculate the significance.
% Due to this result, we used the non-parametric Wilcoxon signed-rank test to calculate the significance.

% # use Wilcoxon signed-rank test (stats.wilcoxon)
% # within-subjects design where each participant rates 2 motion planners for different objects would not be considered independent samples. Instead, it would fall under the category of paired samples or dependent samples.
% # each participant is providing ratings for both motion planners, meaning that the measurements or observations are not independent of each other.
% # key characteristic of paired samples is that the observations or measurements in one condition are related or dependent on the observations or measurements in the other condition. In the context of your study design, the ratings provided by each participant for the two motion planners are paired because they come from the same participant, even though they are for different objects.

% \subsection{Experimental validation}

% The results of the validation study show that ...

\subsection{Impact of Clutter on Legibility}

% # MannwhitneyuResult(statistic=2361.0, pvalue=2.0995819702436923e-19)
% # MannwhitneyuResult(statistic=3642.0, pvalue=1.1989780701860201e-11)
% # MannwhitneyuResult(statistic=4802.5, pvalue=6.641483170755259e-08)
% # MannwhitneyuResult(statistic=3701.0, pvalue=2.2917819462635303e-12)

In order to answer H1, we compared the results of the \textit{uncluttered} and the \textit{cluttered} experiment setup for both sections for the first rank. The statistical analysis results indicate a significant difference between the \textit{uncluttered} and the \textit{cluttered} experiment setups, with p-values $<$ 0.0001. 
This significant difference was observed in both legible motion planners. 

Comparing Fig. \ref{section1_setup1} and \ref{section1}, as well as Fig. \ref{section2_setup1} and \ref{section2}  shows that participants demonstrated higher accuracy in selecting the correct object in the \textit{uncluttered} setup than in the \textit{cluttered} setup. The average distance of the participants' guesses to the correct object is consistently larger in the \textit{cluttered} than in the \textit{uncluttered} setup.

\subsection{Legible Motion Planner Comparison}

In order to answer H2, we compared the results for our proposed legible motion planner and the state-of-the-art legible motion planner.

\subsubsection{Uncluttered Environment}

In the \textit{uncluttered} environment setup, the legible motion planners perform similarly well with \mbox{p-values $>$ 0.05} for both trajectory sections and therefore have no significant difference. Figure 4 shows a visualization of the results for the \textit{uncluttered} environment setup. In both sections of the trajectory, participants seem quite certain which object will be picked. 
% The participants' confidence increased as the robot arm approached the target object, as can be seen by comparing the lower average distance values in \ref{section2_setup1} than in \ref{section1_setup1}.
The lower average distance values in Fig. \ref{section2_setup1} than in Fig. \ref{section1_setup1} indicate that the participants were more confident when the robot arm approached the target object. 
% This observation shows that the participants' confidence was positively influenced by the robot arm's proximity to the target object.

\begin{figure}[t]
    \begin{minipage}{0.91\linewidth}
    \centering  
\subfloat{\includegraphics[width=0.57\textwidth]{Figures/LegendCluttered.png}}    
% \subfloat{\includegraphics[width=0.7\textwidth]{Figures/LegendFig4.png}} 
    \end{minipage} \par\medskip
    \setcounter{subfigure}{0}
    \begin{minipage}{.49\linewidth}
    \centering
    \subfloat[][\textit{Cluttered} setup: Section 1]{\includegraphics[width=1.0\textwidth]{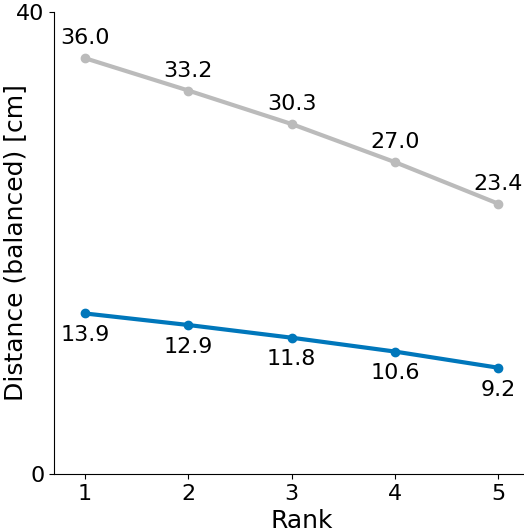}\label{section1}}
    \end{minipage}%
    \hfill
    \begin{minipage}{.49\linewidth}
    \centering
    \subfloat[][\textit{Cluttered} setup: Section 2]{\includegraphics[width=1.0\textwidth]{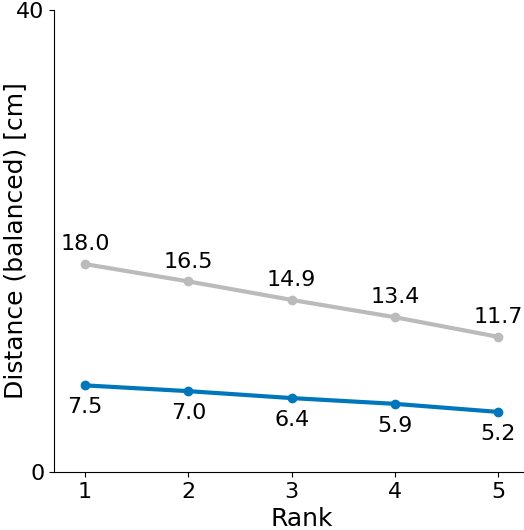}\label{section2}}
    \end{minipage}\par\medskip
    \caption{
    In a \textit{cluttered} environment our entropy-scaled potential field planner performed significantly better compared to the current state-of-the-art legible motion planner with a \protect\subref{section1} \mbox{p-value $<$ 0.0001} and a \protect\subref{section2} \mbox{p-value $<$ 0.05} (calculated with the Wilcoxon signed-rank test for the first rank). Especially in the first trajectory section, participants seemed uncertain regarding which object the robot would grasp (lower distances to the target object are better).
    % In the first trajectory section  \protect\subref{section1}, participants seemed uncertain which object the robot wants to grasp. In the second section \protect\subref{section2}, participants answered on average with a closer distance towards the correct object. 
    % with our entropy-scaled potential field  planner than with the state-of-the-art planner after observing the second section of the trajectory. 
    }
    % \vspace{-1em}
    \label{fig_rank}
  \end{figure}  

\subsubsection{Cluttered Environment}

In the \textit{cluttered} environment setup, our entropy-scaled potential field planner performs significantly better than the current state-of-the-art legible motion planner. This was tested, since we cannot assume normality (see beginning of Section \ref{Results}) with the non-parametric Wilcoxon signed-rank test. For the first section of the trajectory the significance test resulted in a \mbox{p-value $<$ 0.0001} and for the second section of the trajectory the significance test resulted in a \mbox{p-value $<$ 0.05}.

Further, participants answered on average with a closer distance towards the correct object with our entropy-scaled potential field  planner than with the state-of-the-art planner since lower distances to the target object are better. That means that participants are more confident in choosing the correct object and therefore this shows an improvement of our our entropy-scaled potential field  planner compared to the current state-of-the-art legible motion planner.
% The results are marginally significant with a \mbox{p-value $<$ 0.1}, see \mbox{Fig. \ref{fig_rank}}. 
Further, participants are still uncertain after observing the first section of the trajectory which object the robot is reaching for, while in the second section of the trajectory, participants are better at guessing the target object (lower distance of rank 1 in \mbox{Fig. \ref{section2}} than in Fig. \ref{section1}).

\subsection{Open-Ended Questions}

In terms of what would make legible motion planners better for \textit{cluttered} environments, we left the question open to participants of the study and we will include interesting responses as well as our thoughts below.

Overwhelmingly, participants recommended smoother motions in contrast to the potential fields planner and found that any sharp turns to be confusing. Many participants who felt this way also recommended straight lines. While this was found to be less accurate by Bodden, et al., it was not tested in \textit{cluttered} environments so the results of a straight line planner could be worth testing again. Many participants also found the sudden stops of the motion to be confusing so it is worth considering only testing full motions for experimental validation resetting each time a guess is taken. This suggestion only pertains to the overall collection of results, however, and is not necessarily helpful for the design of the motion planner. Some participants recommended more direct motions toward the object rather than the ``hovering" behavior of the state-of-the-art planner. Overall, we did not receive consistent answers to the open ended question, so we also think that personal preference plays a part in how people think the robot should act.

\section{Discussion} \label{Discussion}

\subsection{The Effect of Clutter on Legibility}

The results of the experimental validation study show a significant improvement in legibility of our proposed legible motion planner compared to the current state-of-the-art legible motion planner in \textit{cluttered} environments. The results also emphasize the need to further explore legible motion in \textit{cluttered} environments.
% The results of the experimental validation study show that our proposed entropy-scaled potential field legible motion planner leads to a marginally significant increase in legibility compared to the state-of-the-art legible motion planner. 
We showed that potential fields can be used to induce legibility of robot motion by avoiding to move towards objects other than the target. This is most likely due to the improvements that we made to the optimization criteria presented by Bodden, et al. \cite{bodden2018flexible}, for application in \textit{cluttered} environments.

Based on the experimental results, H1 is \textit{supported}, as shown by the significant differences between the \textit{uncluttered} and the \textit{cluttered} experiment setups, which can be observed by comparing Figure \ref{fig_rank_setup1} and Figure \ref{fig_rank}.

As expected, H2.1 is \textit{not supported} by the experiment results, as shown in the results that are not significantly different between the two planners in Figure \ref{fig_rank_setup1}. Since uncluttered environment trajectories are very similar, the entropy-scaled potential fields based planner used a different strategy to optimize the same parameters as the state-of-the-art-planner.

H2.2 is \textit{supported}, as shown in the results reported in Fig. \ref{fig_rank}. This is due to the fact that there are both more objects and the objects are \textit{cluttered}, which can mean they are bunched together and lowers the legibility of motions towards specific objects. This lowers the potential benefits that a person could receive from collaborating with a robot using either of these legible motion planners. If that person cannot guess reliably what the robot is reaching for in the environment presented in this paper, then in more \textit{cluttered} environments or more sensitive environments they will not feel as comfortable collaborating with and trusting the robot. \par

% It would be dangerous for a person to work with any robot in situations where misunderstanding their intent could compromise their safety. \par

Our proposed entropy-scaled potential field legible motion planner performed significantly better in the \textit{cluttered} environment than the current state-of-the-art legible motion planner. However, to further improve the performance of legible motion planners in \textit{cluttered} environments, it is necessary to conduct more research as indicated by the open-ended participant responses.
% Therefore, it is clear that neither of the motion planners we tested are adequate as a solution to legible motion planning in \textit{cluttered} environments. 
For example, there have been solutions in the past that have been tested and shown to be comparatively worse in \textit{uncluttered} environments; since we have not found a correlation for performance in one environment for predicting performance in the other, we cannot say whether or not those solutions should continue to be discarded as sub-optimal.

\subsection{Recommendations for Legible Motion Planner Research}

% More than anything the results we achieved in this experiment outline three main points: the parameters for which the previous state of the art optimizes are not enough to solve Legible Motion in \textit{cluttered} environments, results for accuracy from experimentation in \textit{uncluttered} environments do not imply similar results in \textit{cluttered} environments, and further research into Legible Motion planning should be performed and validated only in \textit{cluttered} environments.\par

Legible motion is currently focused on overly-controlled environments with a limited number of regularly spaced objects. However, scenarios where robot collaboration is useful are primarily uncontrolled with a variably large number of objects. The goal of research performed in this area is to build systems for implicit communication on the part of the robot which will enable the human to guess its intent in these scenarios. An underlying assumption in prior work is that by studying a simpler version of this problem, \textit{uncluttered} environments, solutions can be created, which will then extend to harder versions of the same problem, \textit{cluttered} environments. Through the results we obtained in this study, we have conclusively shown that this assumption is not true and that there are too many added complexities due to proximity of objects to each other as well as the number of choices of objects.

% In support of the goal of encouraging more research into legible motion in \textit{cluttered} environments, we propose two methods which were tested through the study performed in this research: (1) a novel measure of the \textit{clutteredness} of the environments based on information theory, and (2) a measure of trajectory legibility within an environment. Using this measure, we found a significant correlation as shown in Fig. \ref{correlation_plot} to the accuracy with which people can guess closer to the object consistently.

% \begin{figure}
% \centerline{\includegraphics[width=0.35\textwidth]{Figures/CorrelationPlotR.png}}
% \caption{
% This figure shows the correlation between our proposed entropy-scaled potential field legible motion planner's accuracy and illegibility score for the \textit{cluttered} environment objects of the experimental validation study. Since Spearman's $\rho$ $>$ 0.6 this shows strong correlation. 
% }
% \vspace{-1em}
% \label{correlation_plot}
% \end{figure}

While research into legible motion is still fairly new, we do think that the following suggestions would enable researchers to better collect and to better evaluate their results as research continues into more \textit{cluttered} environments. First and foremost, a benchmark of accuracy and legibility should be studied based on how well people show their intention. We have shown in this paper that the state-of-the-art planner does not perform well in \textit{cluttered} environments and so comparing against it can only give us relative information about how well our planner performed. It would be useful to develop benchmarks for success through accuracy metrics measured in human-human legible motion studies that could then be compared against in robot-human studies. Second, it would be useful to make a delineation between legible motion research where the expectation is that the collaborator is trained to work with the robot or has not worked with the robot in the past. We think that different strategies could be successful in one scenario and not in the other and therefore the delineation would help to evaluate the different strategies more accurately based on their intended use. For instance, an intuitive metric for a trained collaborator would be consistency of approach. However, if the robot is only consistent it is not necessarily helpful for an untrained collaborator. Finally, there may be significant differences between evaluation studies performed in person on a robot and therefore it would be helpful to how this difference can affect the metrics of legibility planners in order to see if a simulated environment is a viable testing stage for this problem. Our particular areas of concern for studies performed in simulation are that the presence of the robot in person is different, the perception of the participant is limited in simulation whereas the perception of a robot in the real world is not, and the field of view for the participant is limited in simulation whereas the participant can move around and take any field of view in the real world.\par

Overall, the results of this validation study demonstrate that clutter significantly impacts the planner's performance. It emphasizes the importance of clutter in testing planners since a planner that performs well in an \textit{uncluttered} environment is not necessarily valid in a \textit{cluttered} environment.

% However, both legible motion planners achieve a low accuracy for cluttered environments. While it is reasonable that legibility of movements in cluttered environments decreases compared to uncluttered environments due to the close proximity of more objects we expect that further research investigating legible motion in cluttered environments is going to increase its accuracy. \par
% The open-ended question ``What would make the motion of the robot more legible?" provided interesting insights into people's preferences regarding legible motion. Several participants mentioned that they would prefer natural movements in the same direction and smoother movements. Also, participants commented that the height, or more precisely, the z-coordinate, has a considerable influence on the legibility of the motion. These legible motion design considerations can be used in future studies to expand on legible motion in cluttered environments. \par
% The experimental validation study also allowed us to construct a legibility measure, see sections \mbox{\ref{LegibilityMeasure}} and \mbox{\ref{QuantitativeValidation}}, which is positively correlated to the accuracy of the experimental validation study, and therefore can be used to improve legible motion. \par
% Further, the experimental validation study results show the need for more realistic study setups since it is more common in the real world that humans work together with a robot in a cluttered environment than in a simpler uncluttered environment. \par

% \section{Limitations and Future Work} \label{Limitations}

\section{Conclusion} \label{Conclusion}
We investigated legible robot motion in \textit{cluttered} environments. We presented a novel definition of the \textit{clutteredness} of objects in a table-top environment and a definition of the legibility with which an object can be grasped. 

Previous legible robot motion planners were tested in \textit{uncluttered} environments. Since \textit{cluttered} environments frequent human-robot collaboration, we tested the previous state-of-the-art and a logical extension of that method based on entropy-scaled potential fields for both \textit{uncluttered} and \textit{cluttered} environments.
The validation study results show a significant improvement in legibility over the state-of-the-art legible motion planner in \textit{cluttered} environments. However, the results also show the need to explore legible motion in \textit{cluttered} environments further to develop an adequate solution to generalizable legible motion planning.
% The validation study results are promising, showing marginally significant increase in legibility over the state-of-the-art legible motion planner. 

The results show that testing legible motion planners in \textit{uncluttered} environments may not necessarily produce outcomes that are applicable to \textit{cluttered} environments. As such, it is imperative to conduct research that takes into account the impact of clutter on the performance of these motion planners. Our recommended modifications to the method of researching legible motion planners are expected to result in more generalizable planners that are applicable in real-world environments.

% Although the results show that testing legible motion planners in \textit{uncluttered} environments yields results which are inconsistent with \textit{cluttered} environments, we believe that our recommended changes to the conduct of research into legible motion will help to create more generalizable planners.

% Altogether, the results show testing Legible Motion planners in \textit{uncluttered} environments yields inconsistent results with those that can be obtained in \textit{cluttered} environments. Therefore, our primary recommendation is that further research in this area is validated in the real world, through a validation study, and with a \textit{cluttered} environment of objects.

%\section*{Acknowledgments}

\bibliographystyle{IEEEtran}
\bibliography{IEEEexample}

\end{document}